\pdfoutput=1

\documentclass[11pt]{article}

\usepackage{acl}

\usepackage{times}
\usepackage{latexsym}

\usepackage[T1]{fontenc}

\usepackage[utf8]{inputenc}

\usepackage{microtype}

\usepackage{inconsolata} 
\usepackage{graphicx}  
\usepackage{subfig}    
\usepackage{float}     
\usepackage{amsmath}
\usepackage{multirow}
\usepackage{pifont}
\usepackage{utfsym}
\usepackage{booktabs}

\usepackage[linesnumbered,ruled,vlined]{algorithm2e}

\newtheorem{example}{Example}           
\newtheorem{definition}{Definition}
\newtheorem{property}{Property}

\newtheorem{proof}{Proof}
\newtheorem{theorem}{Theorem}

\title{Context-Driven Index Trimming: A Data Quality Perspective to Enhancing Precision of RALMs}
\author{Kexin Ma \thanks{These authors contribute equally to this work. } \and Ruochun Jin\footnotemark[1]\thanks{The corresponding author.}  \and Xi Wang \and Huan Chen\and Jing Ren \and Yuhua Tang \\
  Department of Intelligent Data Science, College of Computer Science and Technology, \\National University of Defense Technology, Changsha, China \\
  \texttt{\{makexin, jinrc\}@nudt.edu.cn} \\}

\newcommand{\eat}[1]{}
\newcommand{\revise}[1]{{\color{black}{#1}}}
\newcommand{\warn}[1]{{\color{black}{#1}}}

\begin{document}
\maketitle

\begin{abstract}
\eat{
With the rise of large language models (LLMs), Retrieval-augmented Generation (RAG) has demonstrated strong capabilities in NLP.
Researchers implemented various approaches to enhance the accuracy of retrieval.
However, previous research tend to assume that the tools used for retrieval, vector databases, are ideal and reliable, which is not the case in fact.
We propose a method called \textbf{Context Driven Index Trimming (CDIT)}, modifying the indexes of vector databases and further enhancing the overall quality of generated answers from a more fundamental perspective on data quality.
CDIT logically represents the constraints of data using CMDs (Context Matching Dependency) which are proposed based on MDs (Matching Dependency) and employs LLMs to extract and compare semantic components. 
Furthermore, CDIT identify sentence similarity and trim the indexes of database using CMDs.
Experiments show that models processed by CDIT outperform original models on various tasks.
Moreover, CDIT has been validated to be combined with different language generators, index structures and RAG methods well.
}

Retrieval-Augmented Large Language Models (RALMs) have made significant strides in enhancing the accuracy of generated responses.
However, existing research often overlooks the data quality issues within retrieval results, often caused by inaccurate existing vector-distance-based retrieval methods.
\revise{We propose to boost the precision of RALMs' answers from a data quality perspective through the Context-Driven Index Trimming (CDIT) framework, where Context Matching Dependencies (CMDs) are employed as logical data quality rules to capture and regulate the consistency between retrieved contexts.}
\eat{The essence of CDIT lies in leveraging Context Matching Dependencies (CMDs) to capture and regulate the consistency between retrieved contexts.}
\revise{Based on} the semantic comprehension capabilities of Large Language Models (LLMs), CDIT can effectively identify and discard retrieval results that are inconsistent with the query context and further modify indexes in the database, thereby improving answer quality.
Experiments demonstrate \revise{average improvement of 3.75\% in accuracy}
on challenging question-answering tasks.
Also, The flexibility of CDIT is verified through its compatibility with various language models and indexing methods, \revise{which offers a promising approach to bolster RALMs' data quality and retrieval precision jointly.}

\end{abstract}

\section{Introduction}
\eat{\par Recently, the study of large language models(LLMs) such as the GPT series[Brown et al., 2020, OpenAI, 2023] has become a hot topic in the field of natural language processing(NLP). 
In particular, retrieval-augmented LLMs can effectively address issues such as hallucination \cite{Huang2023ASO}, while also enabling the update of the knowledge required for LLMs with minimal cost and providing explanations for the content generated by LLMs.
Therefore, retrieval-augmented language models have garnered extensive attention from academia.}
\eat{However, not all retrieved citations are useful for the generation result, sometimes retrieval reduces accuracy factually, which has been shown by several recent work \cite{Lewis2020RetrievalAugmentedGF}.}
\eat{Retrieval augmentation may have a negative impact, while irrelevant retrieval may even interfere with the generation result, leading to losses in the result.}

\eat{Therefore, screening the retrieved results after retrieval and selecting to retain those related to the target theme as much as possible is a good method to improve the quality of retrieval-augmented LLMs.}

\revise{Retrieval-augmented large language models (RALMs) have drawn extensive attention, as they effectively ameliorate hallucination \cite{Huang2023ASO}, update the knowledge required for LLMs with minimal cost \cite{Lewis2020RetrievalAugmentedGF}, and provide explanations for contents generated by LLMs \cite{Gao2023RetrievalAugmentedGF}.}
\revise{However, recent study has demonstrated that not all retrieved citations are useful for the generation result \cite{Liu2023LostIT,Wang2023SelfKnowledgeGR}, }where retrieval may reduce the quality of generation.
\revise{For example, if retrieval contains information conflicts, the generation quality may deteriorate,
which leads to false answers to factual questions \cite{Liu2023RECALLAB}.}

\begin{figure}[!t]
    \centering    
    \includegraphics[width=3in]{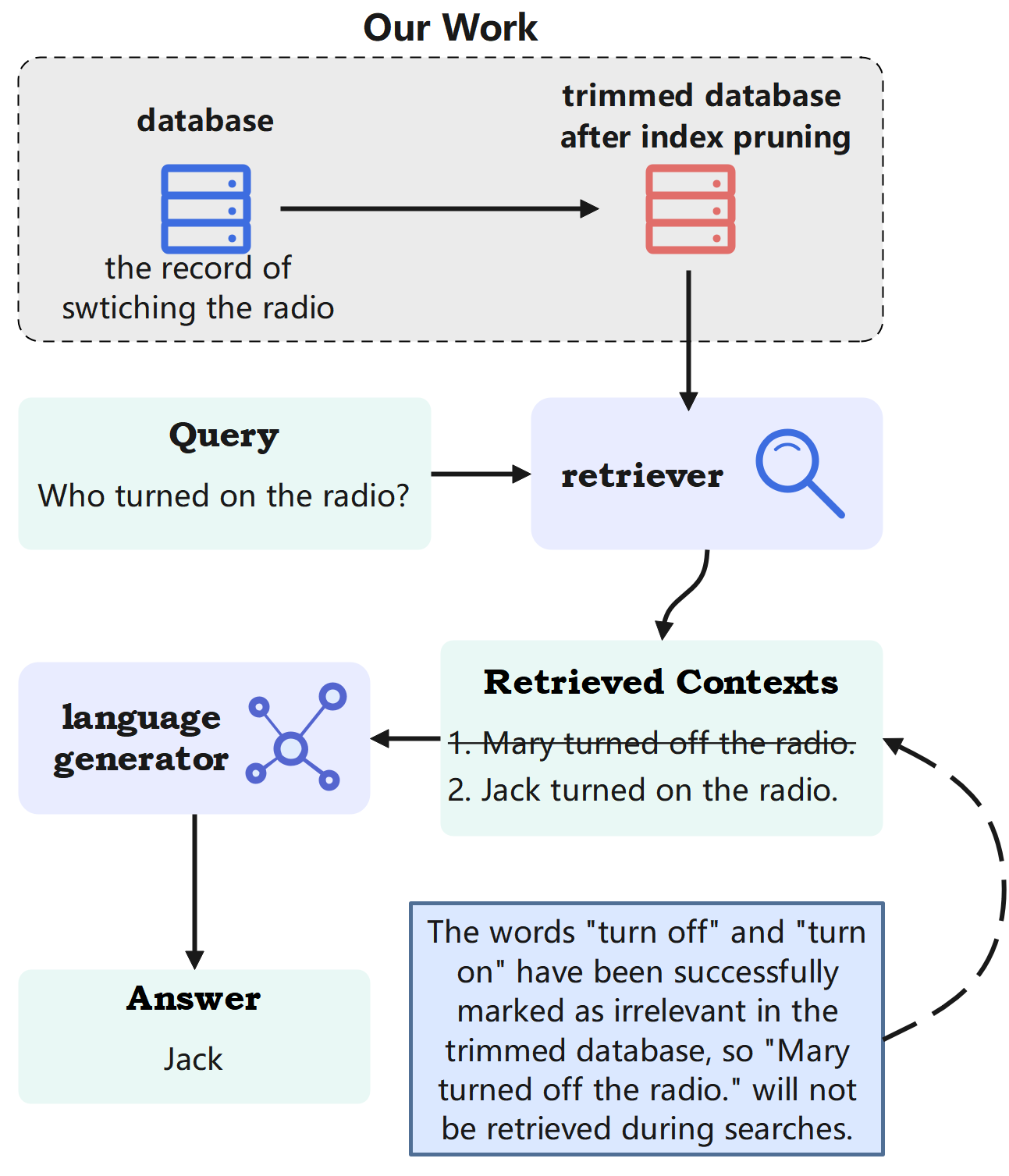}
    \caption{Improve data quality of database to enhance the accuracy of generated answers by RALMs.}
    \label{fig:framework_intro}
    \vspace{-1.2em}
\end{figure}
In view of the challenges \eat{mentioned before}\revise{above}, the NLP community \revise{mainly} focus on enhancing the \revise{retrieval precision}, \eat{such as}\revise{e.g.} ARR \cite{Yu2023AugmentationAdaptedRI}, REPLUG \cite{Shi2023REPLUGRB}, and Atlas \cite{Izacard2022FewshotLW},
\revise{which learns to align the retriever outputs with the preferences of LLMs.}
\revise{However, these works pay little attention to the data quality of knowledge itself.}
\revise{More specifically}, \revise{it is commonly assumed that}\eat{ researchers assume} \revise{data in }the knowledge base (usually implemented as vectors in a vector database) is consistent, \eat{is completely trustworthy, }which is \revise{actually} not the case \revise{in real-world applications}\eat{ in fact}. 
Some sentences appear contextually similar yet are actually opposite in reality. 
\revise{For example,}
consider \revise{two} sentences "He turned on the radio" and "He turned off the radio".
\revise{It is obvious that the semantic meanings of actions in these two sentences are completely opposite.}
\revise{However, the distance between vector representations of the two tends to be small, as most text embedding models are trained to project sequences of tokens that frequently co-occur as neighbouring vectors in the high dimensional semantic space \cite{Li2018WordEF}.}
\revise{Thus, existing vector-distance-based retrieval methods implemented in vector databases may treat these two sentences as ``highly similar'' knowledge, and provide such irrelevant or even conflicting sentences as referring knowledge to the down-stream language generator, which confuses the LLMs and deteriorates the quality of the generated answer\cite{Lewis2020RetrievalAugmentedGF}.}

\par 
\revise{As shown in Figure \ref{fig:framework_intro}, we approach the issue of retrieval quality from the perspective of data quality.
Specifically, inspired by Matching Dependencies(MDs), a classical rule-based data quality management method in the database community \cite{Fan2011DynamicCF}, we propose \textbf{Context Matching Dependencies (CMDs)} that capture and regulate the consistency between the knowledge context and its vector representation.
Then we establish a \textbf{Context-Driven Index Trimming (CDIT)} framework that mainly utilizes CMDs and LLM to improve the quality of RALMs answers by trimming the indexes of vector database.
The CDIT framework starts with an initial retrieval by the retriever in RALMs.
Then the preliminary retrieval results are sent to the CMDs where \warn{an} \eat{a} LLM is employed to determine whether the retrieved knowledge conforms \warn{to} \eat{with} the CMDs constraints.
If the retrieval satisfies the CMDs, it will be passed to LLMs following conventional RALMs.
Otherwise, the retrieval will be discarded, and the vector-search index related to this retrieval will be corrected such that future similar queries can avoid unrelated retrievals return by the vector database.}



\eat{In our process, an initial retrieval is performed by a retriever at first.
Then the preliminary results are sent into the CMDs constructed manually, and are determined whether the constraints are met by a state-of-art LLM.
If the constraints are satisfied, the retrieval result will be preserve and pass on to the next step, otherwise it will be discarded. 
Next, CDIT modify the indexes based on the judgments from LLM. 
If a context was judged to be consistent with a query while another context was not, then we assume the two contexts to be inconsistent. 
Subsequently, we prune the indexes of the two, allowing the vector database to remember that they are dissimilar actually. }

We experimentally verify the effectiveness of CDIT \warn{by open-domain question answering}.
In addition, we \revise{integrate} CDIT with different \warn{language generation models} \eat{such as Llama2-7b}and index construction methods, demonstrating the flexibility of our framework.
CDIT surpasses the basic models with average accuracy improvements of 3.75\% on vairous language models. 
It also boosts the model accuracy by 3.44\%, 4.07\%, and 3.75\%  over IndexFlatL2, IndexHNSWFlat, and IndexIVFFlat, respectively.
Among them, the highest performance improvement can reach up to 15.21\%.

Our main contributions are as follows:
\begin{itemize}
\item \revise{We propose Context Matching Dependencies (CMDs) that maintain consistency among vector data to address the challenge of poor retrieval in RALMs from the perspective of data quality management.}
\item \revise{We develop the Context-Driven Index Trimming (CDIT) framework based on CMDs and LLMs to improve the quality of RALMs answers by trimming the indexes of vector database, which is applicable to any RALM.}
\item \revise{We experimentally verify the effectiveness of CDIT, where \warn{the average and the most significant improvement can reach up to 3.75\% and 15.21\% respectively.}}

\end{itemize}

\section{Related Work}
\revise{\textbf{Retrieval Improvements in RAG.}}
Not all of the retrieved contexts benefit the final results\cite{Liu2023LostIT,Wang2023SelfKnowledgeGR}.
\revise{In order to improve the retrieval quality,
previous work mainly focuses on fine-tuning the retriever to align with the language model.
For example, REPLUG\cite{Shi2023REPLUGRB} freezes the \warn{parameters of language model LLM} and optimizes the retriever to adapt to the language model.
Atlas\cite{Izacard2022FewshotLW}, by contrast, jointly trains the retriever and the language model.}
\revise{Additionally, other work explores improving strategies before and after retrieval.
Specifically, document segmentation strategies \cite{Touvron2023Llama2O} and embedding models \cite{Karpukhin2020DensePR} can be improved before retrieval.
Diversity Ranker in Haystack\cite{Haystack} and LostInTheMiddleRanker\cite{Liu2023LostIT}, on the other hand, investigates document re-ranking after retrieval.}
\revise{Different from previous methods, we approach the retrieval quality issue from the perspective of data quality management,
where the vector index is trimmed based on data consistency captured by CMDs and LLMs.}
\\ \revise{\textbf{Data Quality Rules}.}
\revise{Various logic-based rules and dependencies have been proposed for data quality management.} 
For instance, Functional Dependencies (FDs) \cite{Codd1971FurtherNO} were first introduced in the 1970s to represent integrity constraints and relationships among data.
\revise{Based on FDs, } Conditional Functional Dependencies (CFDs) \cite{Fan2008ConditionalFD} have been proposed for data cleaning purposes.
They use conditions to specify the subset of tuples on which a dependency holds.
Subsequently, Matching Dependencies (MDs) \cite{Fan2011DynamicCF} have been proposed to identify records representing the same real-world entity.
Approximate Functional Dependencies (AFDs) \cite{KaregarGGKSS21} have been proposed to tolerate partial violation tuples to handle noisy datasets better.
In addition, Association Rules (ARs), which were first used to capture item relationships in transaction data, have also been widely studied for data repair and association analysis on relationship data.
Similar rules have been applied on graphs \cite{Galrraga2013AMIEAR,Cao2023ExtractingGP,Fan2022LinkingEA}, to analyze social networks by extracting relations \cite{Erlandsson2016FindingIU,Cagliero2013DiscoveringGA}.
Graph Association Rules (GARs) \cite{Fan2015AssociationRW,Fan2016AddingCQ,Fan2020CapturingAI} have defined association rules directly on graphs, for graph data analysis \cite{Fang2016EffectiveCS,Song2016MiningSF} and knowledge graph search \cite{Namaki2017DiscoveringGT}.
\revise{However, all these data quality rules are designed for relations or graphs, which can hardly support vector data quality management tailored to RALMs.}
\revise{\\ \textbf{Vector Indexing Strategies}.}
Vector databases facilitate efficient similarity search\eat{es} using specialized indexing structures \eat{like}\revise{such as} KD-trees\cite{Bentley1975MultidimensionalBS}, R-trees\cite{Guttman1984RtreesAD}, \revise{and HNSW \cite{Malkov2016EfficientAR}.}\eat{ or more advanced data structures like Hierarchical Navigable Small World (HNSW)\cite{Malkov2016EfficientAR}. }
\revise{In Faiss \cite{douze2024faiss}, there are numerous indexing \eat{methods}\revise{implementations} available, including IndexFlatL2, IndexHNSWFlat, IndexIVFFlat and so on.}
\revise{Although IndexFlatL2 is relatively slow and memory-intensive, it achieves the highest in precision\cite{douze2024faiss}.}
\revise{By contrast, IndexHNSWFlat is fast during searches, at the cost of long index building time and large memory space\cite{Malkov2014ApproximateNN}.}
\revise{Moreover, all these methods are inadequate for RAG since the ANN-based indexing can hardly distinguish statements that are \warn{literally similar} but semantically different\cite{Noonan201552}.}
Several studies have noted the impact of indexing and resorted to re-ranking after retrieval, such as Diversity Ranker in Haystack\cite{Haystack} and LostInTheMiddleRanker\cite{Liu2023LostIT}.
\revise{However, they failed to recognize that \warn{indexing method of the vector database} supporting RAG is inherently unreliable.}

Our research, different from previous studies, attends to \warn{data management of database, modifying the indexing structure to provide a more ideal vector data source for search potentially.}

\section{Context Matching Dependencies}

\revise{We first recall Matching Dependencies (MDs) \cite{Fan2011DynamicCF} before introducing our methods.}
\revise{Given a relational schema $R$ consisting of a set of attributes $attr(R)$,
for each attribute $ A \in attr(R) $, $ dom(A) $ denotes the domain of $A$.}
\revise{Consider an instance $ r $ of $ R$ and a tuple $t\in r$, then for $ \forall A\in attr(R), t[A]\in dom(A)$, where $t[A]$ represents the projection of $t$ onto $A$.}
\revise{Matching Dependencies (MDs)} are defined to match the attributes of different tuples as follows.

\begin{definition}
Matching Dependency
$$
    \bigwedge_{j\in [1,k]}(r_1[A_j] \approx_j r_2[B_j]) \rightarrow \bigwedge_{i\in [1,h]}(r_1[E_i] \rightleftharpoons r_2[F_i])
$$
\end{definition}

\revise{where for $ \forall j\in [1,k], \forall i\in [1,h]$, $A_j$ and $E_i$ are attributes of $r_1$, $B_j$ and $F_i$ are attributes of $r_2$,
$\approx$ is the similarity predicate which returns true if the two attributes are regarded as similar, $\rightleftharpoons$ is the matching operator which indicates that the attributes are \warn{identified} \cite{Fan2011DynamicCF}.}


\eat{where for $ \forall j\in [1,k], \forall i\in [1,h]$, $A_j$ and $E_i$ are attributes of $r_1$, $B_j$ and $F_i$ are attributes of $r_2$.
In addition, $\approx$ is a similarity predicate that denotes the attributes can be regard as similar in matching, while $\rightleftharpoons$ denotes a matching operator, indicating that attributes are identified.}

\eat{Our work primarily serves for RAG, therefore we specially defined Context Matching Dependency (CMD) for vector database using the form of MD for reference.
$\sim$ denote that the attributes are consistent.}

\revise{Similar to relational database tuples that consist of multiple attributes,
natural language sentences can be represented by their linguistic components such as subject, predicate and object\cite{Stefanescu2014ASS}.}
\revise{Thus, inspired by MDs for entity resolution in relational databases,
we can determine whether two sentences are semantically similar
based on the similarity between their corresponding linguistic components.}
\revise{Let $sub$, $pre$ and $obj$ denote the subject, predicate and object of a sentence, respectively.
We also define \textit{semantic id} ($sid$) that denote the semantic meaning of a natural sentence.
Similar to "id" as the main key in relational databases, $sid$ identifies a sentence in the high-dimensional semantic space.
If sentences $s_1$ and $s_2$ have similar semantics,
then their semantic ids are consistent, denoted as $s_1[sid] \sim s_2[sid]$.}

\eat{Now take a holistic view of the problem facing currently.
Natural sentences have different semantic meanings which can be represented by their components to a certain extent.
Therefore, to determine whether the two sentences are similar, the linguistic features of natural sentences are used.}

\eat{We use $sub,pre,obj$ to denote the subject, predicate and object of a sentence.}
\eat{Meanwhile, a component named \textit{semantic id} ($sid$) is defined as a similarity measurement,
denoting the actual semantic meaning of a natural sentence in real-life situations.
Similar to the concept of "id" in relational databases, $sid$ identifies a sentence in the semantic space. 
If the $sid$s of two sentences are judged to be the consistent, then we assume they have the similar semantics.
To explain it more specific, we give an example.}

\begin{figure}[!ht]
    \centering 
    \includegraphics[width=3in]{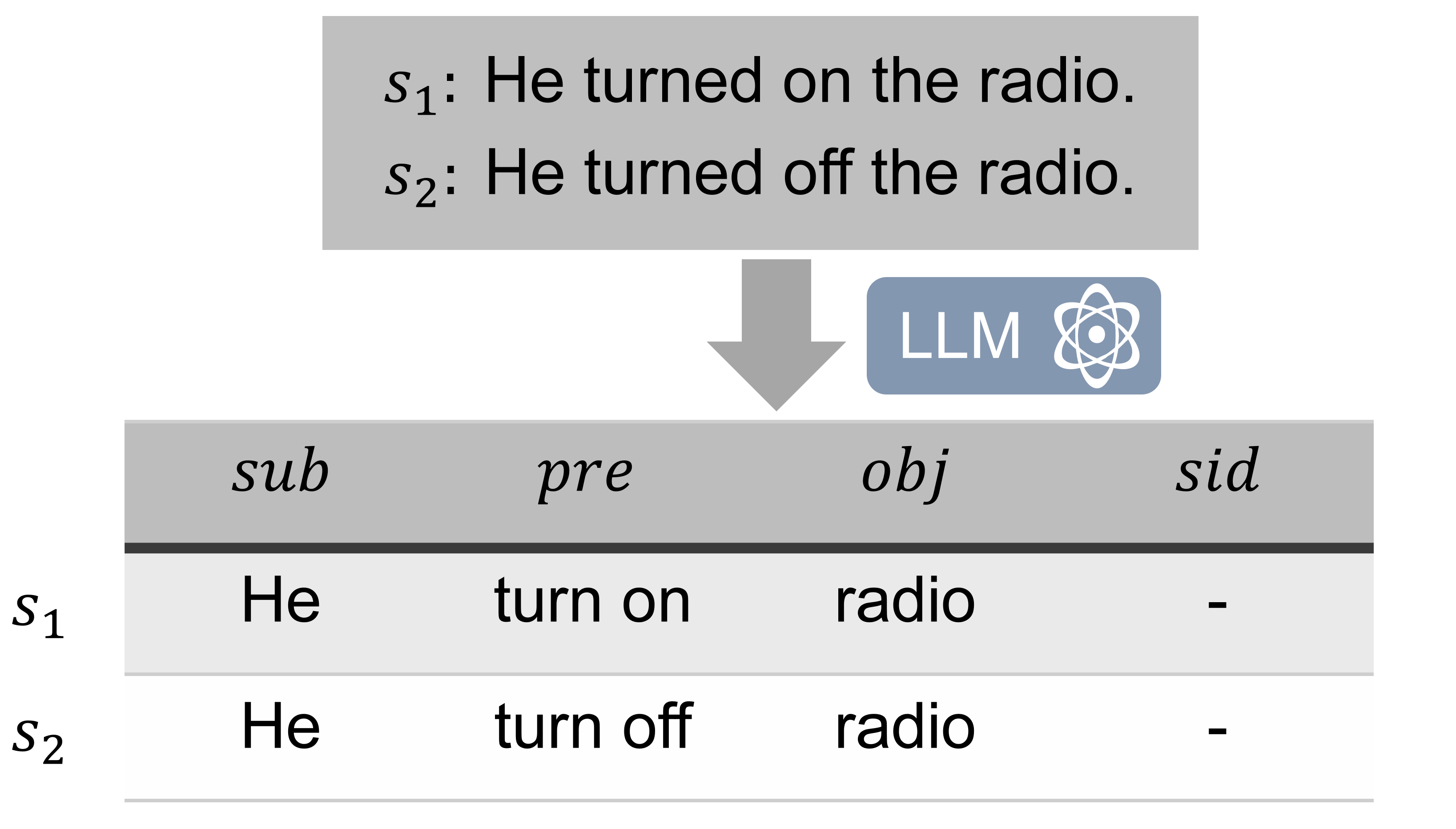}
    \caption{Example relational data of natural language}
    \label{fig:tab}
\end{figure}

\begin{example}
\label{ex:ex2}
Consider the \revise{following two sentences}.
\\$s_1:$ He turned on the radio.
\\$s_2: $ He turned off the radio.
\\ \revise{As shown in Figure \ref{fig:tab}, $s_1[sub]$,$s_1[pre]$ and $s_1[obj]$ represents
\textit{He}, \textit{turn on} and \textit{radio} in sentence $s_1$, respectively,
while $s_2[sub]$,$s_2[pre]$ and $s_2[obj]$ denotes
\textit{He}, \textit{turn off} and \textit{radio} in $s_2$, respectively.}
\end{example}

\revise{Inspired by MDs designed for entity resolution in databases,
in order to serve RAG better,
we propose Context Matching Dependency (CMD) to manage data consistency in vector databases.}

\begin{definition}
\label{def:CMD}
Context Matching Dependency 
$$
    (s_1[sid] \sim s_2[sid]) \rightarrow \bigwedge_{i\in [1,k]}(s_1[X_i] \approx_i s_2[Y_i])
$$
\end{definition}
\revise{where $\approx$ means that the corresponding sentence components are similar, $X_i,Y_i$ denote grammatical components in $s_1,s_2$.
In addition, $\not\sim$ and $\not\approx$ denote $sid$ inconsistency and dissimilarity of sentence components, respectively.}
\eat{in Definition \ref{def:CMD}, we actually have:
$$
\bigwedge_{j\in [1,k]}(r_1[A_j] \approx_j r_2[B_j]) \rightarrow \bigwedge_{i\in [1,h]}(r_1[E_i] \rightleftharpoons r_2[F_i])
$$
$$
\bigwedge_{i\in [1,h]}(r_1[E_i] \rightleftharpoons r_2[F_i])  \rightarrow  \bigwedge_{j\in [1,k]}(r_1[A_j] \approx_j r_2[B_j])
$$}

\begin{example}
\label{ex:exemplary-CMD}
Consider the following CMD.
\begin{equation*}
\label{eq:md_r}
    \begin{aligned}
    \phi_1:
&s_1[sid] \sim s_2[sid] \rightarrow s_1[sub] \approx s_2[sub] \land  \\
&s_1[pre] \approx s_2[pre] \land s_1[obj] \approx s_2[obj]
\end{aligned}
\end{equation*}
\revise{$\phi_1$ claims that if the semantic meanings of $s_1$ and $s_2$ are consistent,
then their corresponding subjects, predicates, and objects should be similar in semantics.
Similar to FDs and MDs \cite{Codd1971FurtherNO,Fan2011DynamicCF}, CMD can be applied to vector databases to check data consistency.}
\end{example}

\eat{
For two sentences $s_1,s_2$, we could get rules as below.
\\$\varphi_1$: Only if $s_1$[$sub$] $\approx$ $s_2$[$sub$], $s_1$[$pre$] $\approx$ $s_2$[$pre$] and $s_1$[$obj$] $\approx$ $s_2$[$obj$], $s_1$[$sid$] and $s_2$[$sid$] will be consistent. 

That is to say, \warn{if two sentences are similar, then the subjects, verbs and objects can be identified.}
By rewriting the rules as CMDs, we can obtain the following CMD \ref{eq:md_r}.
\begin{equation}
\label{eq:md_r}
    \begin{aligned}
&s_1[sid] \sim s_2[sid] \rightarrow s_1[sub] \approx s_2[sub] \land  \\
&s_1[verb] \approx s_2[verb] \land s_1[obj] \approx s_2[obj]
\end{aligned} 
\end{equation}
}
\revise{Table \ref{tab:not} summarizes symbols and notations.}

\eat{
According to the CMDs proposed, we can summarize two properties of $sid$: \textbf{symmetry} and \textbf{transitivity}.
Specific proofs of the properties can be found in Appendix~\ref{sec:ap-proof}.
}

\begin{table}[t]
\centering
\scalebox{0.9}{
\begin{tabular}{lc}
\hline
\textbf{symbols} & \textbf{notations} \\
\hline
$R,r$ & relational schema, instance \\
$A,B,E,F$ & attribute \\
$X,Y$ & component \\
$s,s_1,s_2$ & natural sentences\\
$s[X]$ & the corresponding words in sentence \\
$\sim,\not\sim $ & consistency predicate\\
$\approx,\not\approx$ & similarity predicate\\
$\rightleftharpoons$ & the matching operator\\
\hline
\end{tabular}}
\caption{Summary of main symbols and notations}
\label{tab:not}   
\end{table}

\section{Context Driven Index Trimming}
\revise{\subsection{Method Overview}}
\begin{figure*}[!ht]
    \centering    
    \includegraphics[width=6in]{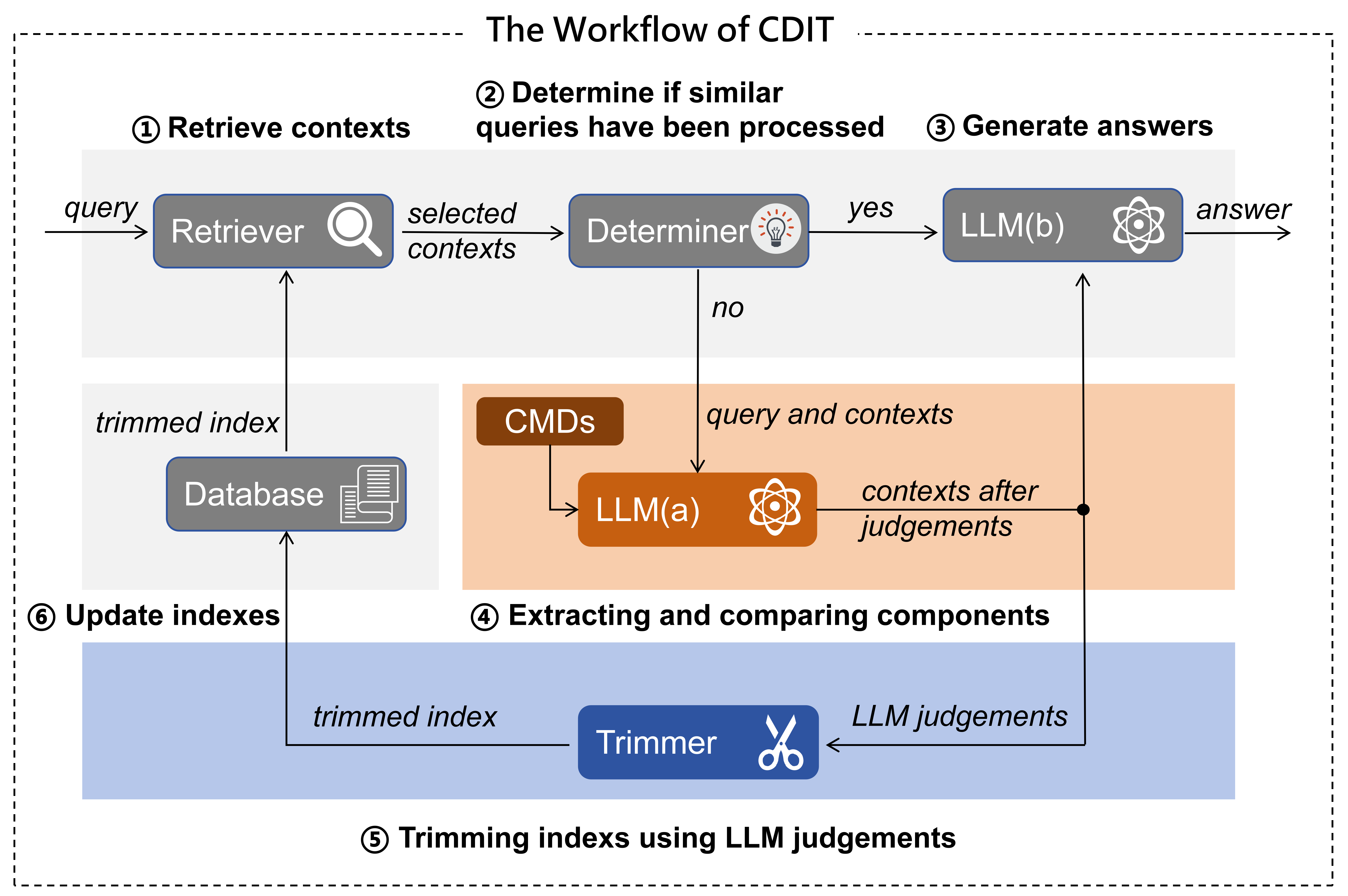}
    \caption{Overview of our mechanism. The LLM(a) represents the more advanced large-parameter language models currently, such as GPT-3.5-turbo; LLM(b) stands for LLMs with smaller parameters and easier deployment, such as Llama2-7b, playing the role of a language generator.}
    \label{fig:framework}
\end{figure*}

\revise{
As shown in Figure \ref{fig:framework}, CDIT starts with an initial retrieval (step \textcircled{\small{1}})
and the query will be checked that whether a similar query has been processed before (step \textcircled{\small{2}}).
If a similar query is found by the determiner via semantic similarity search \cite{Gao2023RetrievalAugmentedGF},
the initial retrieval along with the query will be used to generate the final answer (step \textcircled{\small{3}}).
Otherwise, CDIT employs an LLM to extract the main semantic components of the retrieved sentences and
checks whether the retrieval data and the query conform with the CMDs (step \textcircled{\small{4}}).
Retrieval results that are determined as consistent by the CMDs and the LLM will be preserved and passed to the following steps,
while inconsistent results are discarded.
Later in step \textcircled{\small{5}}, CDIT trims the vector index based on the LLM judgements,
which enables the database to update its vector search index for better retrieval in the future.}
Key steps \textcircled{\small{4}}\textcircled{\small{5}} will be introduced in following sections.



\eat{The overall framework is shown in Figure \ref{fig:framework}.
At first an initial retrieval is performed by a retriever and the query will be checked if a similar one has been processed by CDIT before.
If yes, then the retrieval will use the trimmed index directly with the corresponding query to generate final answers.
This process corresponds to step \textcircled{\small{1}}\textcircled{\small{2}}\textcircled{\small{3}} in the figure.
If no, then we extract the main semantic components of the sentences and compare them in step \textcircled{\small{4}}.
In this step, CMDs determine whether the constraints are met and $sid$s are consistent.
If the $sid$s are consistent, the retrieval result will be preserved and passed on to the next step; otherwise, it will be discarded. 

After that, in step \textcircled{\small{5}} CDIT trim the indexes based on the LLM judgments, allowing the vector database to remember the new relation of data.
Finally, the database updates its indexes using trimmed results in step \textcircled{\small{6}} and sends new indexes to the retriever when retrieving.}


\subsection{Extracting and Comparing Components}
\revise{An LLM is employed to extract and compare the subjects, predicates and objects of sentences
and further judge whether the retrieval data and the query are consistent based on CMDs.}
\revise{Specifically, a prompt consisting of rules and instructions is designed for extraction, comparison and judgement
(see Appendix~\ref{sec:ap-e&c} for details).}
\revise{In the rule part of the prompt, \warn{we explain the meaning of CMDs to the LLM via natural language.}
In the instruction part, we ask the LLM to extract and compare the sentence components based on the CMD and decide whether the data is consistent.}
\revise{In our experiments, we adopt GPT-3.5-turbo as the extraction and comparison model,
which provides accurate judgments and is easy to implement with good flexibility and reasonable price.}
\revise{Continue with Example \ref{ex:ex2}, as shown in Figure \ref{fig:tab},
basic semantic components $sub, pre, obj$ of $s_1, s_2$ are firstly extracted by GPT-3.5-turbo.
After comparison, the LLM finds that $turn\ on$ and $turn\ off$ are dissimilar,
which is denoted as $s_1[pre]\not\approx s_2[pre]$.
Therefore, the CMD $\phi_1$ is violated,
and the LLM returns "False" which means that $s_1[sid] \not\sim s_2[sid]$.}

In summary, the consistency of retrieved contexts and queries are checked in this step,
\revise{where }the LLM decomposes sentences into main components and serves as a comparator \revise{for matching.}
\eat{during the matching process.}

\subsection{Trimming indexes}
\eat{After distinguishing the similarity between queries and contexts, we compare and modify the contexts themselves, pruning the indexes such that data that should not be considered similar in the database can be pushed apart in vector space.}

\revise{We propose an index trimming algorithm (Algorithm \ref{alg:it}) based on the \textbf{Witness Theorem}
to prune incorrect indexes of the retrieved data,
such that inconsistent contexts along with their corresponding vectors no longer link together.}

\eat{We proposed a \textbf{Witness Theorem} \ref{thm:witness} and an algorithm \ref{alg:it} to assist in pruning the indexes between retrieved data, such that data that should not be considered similar in the database can be pushed apart in vector space.}

\eat{\textbf{Witness Theorem} elaborated in Theorem \ref{thm:witness} is proposed to identify the relations that are wrongly considered similar in the original database.}

\revise{\textbf{Witness Theorem} identifies the contexts wrongly considered similar in the vector database.}

\begin{theorem}
\label{thm:witness} 
    Witness Theorem.
    Given a query $q$ and sentences $s_1, s_2$. 
    If $q[sid]\allowbreak\sim\allowbreak s_1[sid]$ and $q[sid] \allowbreak \not \allowbreak \sim s_2[sid]$, 
    then the query $q$ is a \textbf{witness} to \warn{the separation of the two sentences $s1$ and $s_2$.}
\end{theorem}

In other words, if the $sid$ consistency judgement of $q,s_1$ and $q,s_2$ differs, then $q$ witnesses the contradiction between $s_1$ and $s_2$.
\revise{As a sufficient number of witnesses are collected,
it can be determined that $s_1$ and $s_2$ are actually dissimilar.}
In that case, we modify the vector index
by cutting the similarity linkage between $s_1$ and $s_2$.
Algorithm \ref{alg:it} shows this process of trimming indexes.

\begin{figure*}[!ht]
    \centering
    \includegraphics[width=6in]{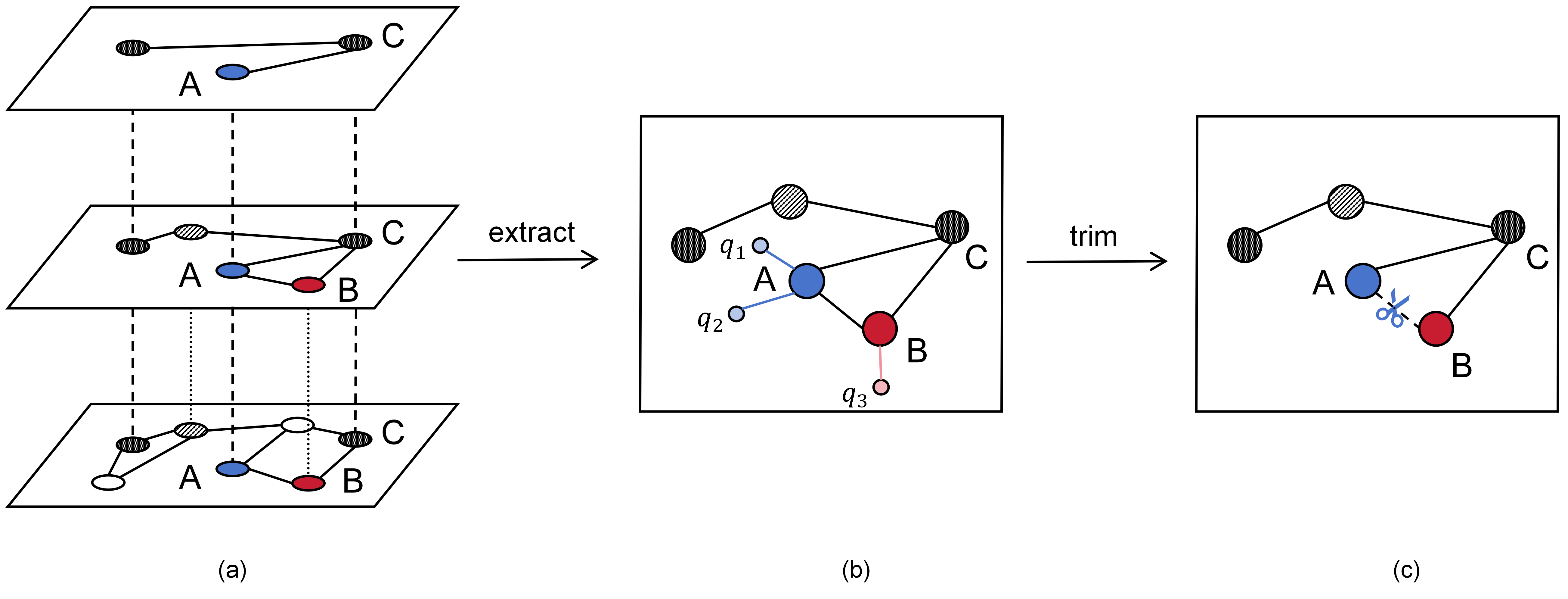}
    \caption{Diagram of HNSW indexing. A, B, and C denote the data vector and $q_1, q_2, q_3$ denote the query vector. (a) The diagram of HNSW structure. (b) A single-layer graph is extracted from the stereoscopic structure in (a). (c)After trimming the indexes, the relationship pointed to by the dashed line was successfully deleted.}
    \label{fig:hnsw}
\end{figure*}

\revise{
To illustrate, we take IndexHNSWFlat in Faiss as an example.
\revise{As shown in Figure \ref{fig:hnsw}, IndexHNSWFlat establishes a vector search index based on HNSW algorithm \cite{Malkov2016EfficientAR},
where data in the knowledge base is organized as a hierarchical similarity graph to facilitate efficient searching.}
From Figure \ref{fig:hnsw}(a), we can see that A and B are regarded as similar and connected by HNSW.
However, as the LLM determines that $q_1, q_2$ are similar to A while dissimilar to B,
and $q_3$ is similar to B but not A (Figure \ref{fig:hnsw}(b)),
there has been adequate number of witnesses for A and B to separate.
Thus, CDIT will cut the edge between A and B (the dashed line in Figure \ref{fig:hnsw}(c))
and the vector search index is trimmed.}

\eat{
Assume a query $q$ and two contexts $s_1,s_2$ retrieved from the corpus based on the query.
Due to the search method based on IndexHNSWFlat, sentences retrieved like $s_1$,$s_2$ are considered similar in the index structure of vector database itself.
According to the Property \ref{thm:transitivity}, we can conclude that:
\begin{align*}
      &q[sid] \rightleftharpoons s_1[sid] \land  q[sid] \rightleftharpoons s_2[sid]\\
      &\rightarrow s_1[sid] \rightleftharpoons s_2[sid]
\end{align*}
This means if $q$ and $s_1$, $q$ and $s_2$ are both identified as similar, then the relation between $s_1$ and $s_2$ should be kept.
}

\begin{algorithm}[!ht]
	\caption{Index trimming}
    \label{alg:it}
	\KwIn{Query}
	\For{q in Query}{
            $T\leftarrow$new empty collection\;
            $s_C\leftarrow$new empty collection\;
		$C\leftarrow$ \textbf{Retrieve}$(q)$\;
            \tcp{Retrieve contexts similar with q.}
            add $C$ to $s_C$\;
            \For{s in C}{
                $result\leftarrow$\textbf{Judge}($s,q$)\;
                \tcp{Judge $sid$s with CMDs.}
                \eIf{result=True}{
			add $s$ to $T$\;
                \tcp{Record similar sentences.}
		      }
                {
                \textbf{Accumulate($s,T$)}\; 
                \tcp{Accumulate the number of witnesses.}
                }
            }	
	}
        \For{$s_1$, $s_2$ in $s_C$}{
        \If{\textbf{Witness}($s_1,s_2$)}{
                \textbf{Cutoff}($s_1,s_2$)\;
                \tcp{If the contexts have been witnessed a certain number of times, then cut off the link for $s_1$ and $s_2$.}
                }
        }
\end{algorithm}

\revise{Although the retrieved contexts are all considered similar by the vector search index,
Witness Theorem assists in pruning incorrect similarity links of the retrieval in the search index.
This way, the next time a similar query is encountered,
the retrieved context will have better data consistency since the index has been previously modified. }

\eat{
when retrieving from the corpus, the query will be projected into the vector space.   
The retriever will first find the context vector closest in distance with the query vector, and then examine the other points adjacent to this vector.

Therefore, take Figure \ref{fig:hnsw} as an example, when querying vector Q, we first found the closest vector A to it. 
Since B and C is linked with A, finally the retrieved contexts will contain A,B and C based on the origin indexes.

However, when compare Q with B, we found that the result is dissimilar, contradicting with the condition mentioned before.
That is to say, Q witnessed that A and B actually have some inconsistency.
When such Q appears frequently, we believe that actually B and A are not similar.
B would appear as an error in retrieved contexts.
Therefore, CDIT cut off the linkage of A and B.
As a result, B will not be retrieved and this improve the accuracy of retrieval.
}

\section{Experiments}

\renewcommand{\arraystretch}{1.0}
\begin{table*}[!ht]
\centering
\scalebox{0.85}{
\begin{tabular}{l|l|ccccccccc}
\toprule
\multirow{2}{*}{\textbf{Basic Model}} & \multirow{2}{*}{\textbf{Index}}  & \multicolumn{2}{c}{\textbf{PopQA}(acc)}  &  \multicolumn{2}{c}{\textbf{TQA}(acc)} &  \multicolumn{2}{c}{\textbf{ARC}(acc)} &\multicolumn{2}{c}{\textbf{Pub}(acc)} & \multirow{2}{*}{\textbf{Avg-Impro}} \\
&  &orignal  &CDIT &orignal &CDIT  &orignal &CDIT &orignal &CDIT \\
\midrule
\multirow{3}{*}{Llama-7b} & IndexFlatL2  &14.35  &\textbf{19.02} &23.82 &\textbf{32.48} &27.47 &\textbf{32.94} &23.10 &\textbf{27.03} & \textcolor{blue}{$\uparrow$ 5.69}\\
&IndexHNSWFlat  &20.85  &\textbf{23.94} &26.48 &\textbf{32.48}  &27.39 &\textbf{30.97} &25.53 &\textbf{29.10} & \textcolor{blue}{$\uparrow$ 4.06}\\
&IndexIVFFlat  &20.43  &\textbf{24.44}  &26.74 &\textbf{32.90}  &32.93 &\textbf{34.22} &23.71 &\textbf{26.94} & \textcolor{blue}{$\uparrow$ 3.67}\\
\midrule
\multirow{3}{*}{Llama2-7b} & IndexFlatL2  &15.76  &\textbf{17.85} &22.61 &\textbf{30.52} &27.47 &\textbf{29.35} &24.51 &\textbf{26.24} & \textcolor{blue}{$\uparrow$ 3.40} \\
&IndexHNSWFlat  &19.52  &\textbf{27.61} &24.49 &\textbf{32.33}  &28.16 &\textbf{29.27} &23.30 &\textbf{26.13} & \textcolor{blue}{$\uparrow$ 4.97}\\
&IndexIVFFlat  &20.26  &\textbf{25.44}  &26.45 &\textbf{30.78}  &30.80 &\textbf{31.06} &26.34 &\textbf{28.74} & \textcolor{blue}{$\uparrow$ 3.05}\\
\midrule
\multirow{3}{*}{Alpaca-7b} & IndexFlatL2  &21.85  &\textbf{27.19} &33.57 &\textbf{43.40} &26.45 &\textbf{27.99} &56.34 &\textbf{60.56} & \textcolor{blue}{$\uparrow$ 5.23}\\
&IndexHNSWFlat  &30.44  &\textbf{35.20} &37.84 &\textbf{47.32}  &28.16 &\textbf{30.46} &56.53 &\textbf{63.26} & \textcolor{blue}{$\uparrow$ 5.82}\\
&IndexIVFFlat  &29.85  &\textbf{34.02}  &37.57 &\textbf{44.14}  &27.05 &\textbf{32.00} &56.53 &\textbf{63.21} & \textcolor{blue}{$\uparrow$ 5.59}\\
\midrule
\multirow{3}{*}{Llama3-8b} & IndexFlatL2  &\textbf{42.00}  &41.87 &39.03 &\textbf{39.66} &\textbf{33.02} &32.08  &46.30 &\textbf{50.11} & \textcolor{blue}{$\uparrow$ 0.80}\\
&IndexHNSWFlat  &41.79  &\textbf{41.87} &38.99 &\textbf{39.45}  &31.22 &\textbf{32.00} &43.26 &\textbf{48.41} & \textcolor{blue}{$\uparrow$ 1.62}\\
&IndexIVFFlat  &41.37  &\textbf{42.20}  &38.23 &\textbf{39.15}  &31.31 &\textbf{33.53} &44.78 &\textbf{49.02} & \textcolor{blue}{$\uparrow$ 2.05}\\
\midrule
\multirow{3}{*}{Mistral-7b} & IndexFlatL2  &40.12  &\textbf{42.70} &60.21 &\textbf{62.41} &55.29 &\textbf{57.25}  &21.48 &\textbf{23.46} &\textcolor{blue}{$\uparrow$ 2.18}\\
&IndexHNSWFlat  &31.11  &\textbf{34.28} &57.58 &\textbf{60.48}  &53.13 &\textbf{56.74} &21.58 &\textbf{24.23} & \textcolor{blue}{$\uparrow$ 3.08}\\
&IndexIVFFlat  &35.45  &\textbf{41.87}  &61.09 &\textbf{62.59}  &54.93 &\textbf{56.48} &23.10  &\textbf{29.21} & \textcolor{blue}{$\uparrow$ 3.90}\\
\midrule
\multirow{3}{*}{Bloomz-7b1} & IndexFlatL2  &24.86  &\textbf{27.52} &49.71 &\textbf{51.14} &40.10 &\textbf{48.89} &56.84 &\textbf{57.98}  &\textcolor{blue}{$\uparrow$ 3.51}\\
&IndexHNSWFlat  &23.52  &\textbf{24.10} &47.71 &\textbf{49.73}  &43.51 &\textbf{48.55} &53.19 &\textbf{55.26} & \textcolor{blue}{$\uparrow$ 2.43}\\
&IndexIVFFlat  &24.19  &\textbf{29.53}  &48.37 &\textbf{53.29}  &43.60 &\textbf{48.63} &57.34 &\textbf{59.21} & \textcolor{blue}{$\uparrow$ 4.30}\\
\midrule
\multirow{3}{*}{Falcon-7b} & IndexFlatL2  &28.69  &\textbf{32.53} &40.02 &\textbf{45.89} &20.04 &\textbf{21.08} &25.32 &\textbf{27.68}  &\textcolor{blue}{$\uparrow$ 3.28}\\
&IndexHNSWFlat  &20.60  &\textbf{25.35} &27.47 &\textbf{42.68}  &21.50 &\textbf{25.26} &23.91 &\textbf{26.21} & \textcolor{blue}{$\uparrow$ 6.51}\\
&IndexIVFFlat  &22.52  &\textbf{29.94}  &37.84 &\textbf{41.71}  &20.01 &\textbf{21.59} &27.56 &\textbf{29.54} & \textcolor{blue}{$\uparrow$ 3.71}\\

\bottomrule
\end{tabular}
}
\caption{Experiment results on different language models and index structure. Bold numbers indicate the best performance among models. "Avg-Impro" refers to the average improvement of CDIT of all types of datasets. \textbf{PopQA,TQA,ARC and Pub} refer to  PopQA, TriviaQA-unfiltered, ARC-Challenge, and PubHealth, respectively.}
\label{tab:result}   
\end{table*}

\subsection{Experimental settings}
\textbf{Datasets.}
\revise{We verify the effectiveness of CDIT on four datasets,
including ARC-Challenge, PubHealth, PopQA and TriviaQA-unfiltered,
which are well-accepted and challenging factual question benchmarks for RAG\cite{asai2023selfrag,Gao2023RetrievalAugmentedGF}.}
\revise{Specifically, \textbf{ARC-Challenge}\cite{Clark2018ThinkYH} is a multiple-choice reasoning dataset
that requires far more powerful knowledge than previous tasks.}
\revise{It contains challenging questions that most retrieval-based algorithms can hardly answer correctly.}
\textbf{PubHealth} is a fact-checking task about public health, containing 987 non-disputed factual and faked claims for evaluating the fact-
check performance.
\revise{For these two datasets, accuracy are the evaluation metrics, calculated by the given ground-truth answers.}
\textbf{PopQA}\cite{Mallen2022WhenNT} contains QA pairs whose questions are generated by converting a knowledge tuple (subject\_entity,object\_entity, relationship\_type) retrieved from Wikidata.
\textbf{TriviaQA-unfiltered}\cite{Joshi2017TriviaQAAL} has complex and compositional questions, raising the need for more precise retrieval.
\revise{We follow the rough matching in \cite{asai2023selfrag,Mallen2022WhenNT} as the performance metric,
where a generation is correct when the ground-truth answer is included.}
\\ \textbf{Configurations in CDIT.}
We employ Contriever-MS MARCO\cite{Izacard2021UnsupervisedDI} as the retriever and Faiss\cite{douze2024faiss} as the vector search interface.
By default, 
\revise{The top-10 documents returned by the retriever are selected in CDIT by default.}
\warn{The official April 2018 English Wikipedia dump is used as the knowledge base. }
\revise{GPT-3.5-turbo is employed to extract and judge the consistency of \warn{$sid$s}.}
CMD rules $\phi_1$ are used as the constraints.
More details can be found in Appendix~\ref{sec:ap-ip}.

\subsection{Baselines}
\revise{\textbf{Language models.}}
We have tested CDIT on various baseline language models that serve as the answer generator in the RAG stage,
including Llama-7b\cite{Touvron2023LLaMAOA}, Llama2-7b\cite{Touvron2023Llama2O}, Alpaca-7b\cite{Dubois2023AlpacaFarmAS}, Llama3-8b\footnote{\url{https://github.com/meta-llama/llama3}}, Mistral-7b\cite{jiang2023mistral}, Bloomz-7b1\cite{muennighoff2022crosslingual} and Falcon-7b\cite{almazrouei2023falcon} in consideration of their convenience, accessibility, and versatility.
\warn{Bloomz is a Multitask Prompting Fine Tuned (MTF) version of the BLOOM\cite{le2023bloom}, and Alpaca is replicated based on Llama, and the instruction-tuned LM adopts an official system prompt during training.}
\revise{\\ \textbf{Vector search indexes.}}
\revise{We have also tested CDIT on various representative vector similarity search indexes including IndexFlatL2, IndexHNSWFlat and IndexIVFFlat\cite{douze2024faiss}.}
\revise{Specifically, IndexFlatL2 performs Euclidean distance search on \warn{all} vectors, which is the most accurate but slow in search and memory-intensive.}
\revise{IndexHNSWFlat is built on the Navigable Small World (NSW) graph, which provides extremely fast search at the cost of both long building time and large memory space for the index.}
\revise{IndexIVFFlat reduces the search space via clustering, which strikes a balance between search quality and speed.}
Unless otherwise specified, the default configuration for the experiment shall be IndexL2Flat.

\subsection{Main Results}
\revise{Table \ref{tab:result} compares the answer accuracy of the original RAG and CDIT, where we find the following:}
\\ \textbf{CDIT works \eat{well in}\revise{for} various language models.}
\revise{CDIT surpasses the basic models with average accuracy improvements
of 4.47\%, 3.80\%, 5.54\%, 1.49\%, 3.05\%, 3.41\% and 4.50\% on Llama-7b, Llama2-7b, Alpaca-7b, Llama3-8b, Mistral-7b, Bloomz-7b1 and Falcon-7b, respectively,
and the most significant increase reaches up to 15.21\%
when applying CDIT framework to \warn{Falcon-7b model} on \warn{TriviaQA dataset} with IndexHNSWFlat index.
This is because retrieval information that is unrelated or inconsistent with the query is discarded by CDIT, which reduces the distracting inputs to LLMs.}
\\ \textbf{CDIT works \eat{well in}\revise{for} different indexing methods.}
\revise{CDIT has on average boosted the model accuracy by 3.44\%, 4.07\%, and 3.75\% over IndexFlatL2, IndexHNSWFlat, and IndexIVFFlat, respectively,
which proves its effectiveness in modifying the original vector index.}
\revise{Moreover, CDIT achieves higher improvements on more coarse-grained indexing structures, such as IndexHNSWFlat.}

\eat{Three basic index methods are utilized in the test.
Results showed that CDIT has respectively boosted the average accuracy of the model by 3.99\%, 3.96\%, and 3.41\% across three index structures, proved effectiveness by reasonably modified the original structures.
What's more, we can see that CDIT has a higher improvement on index structure that were originally less effective such as IndexL2Flat.}

\subsection{Analysis}
\begin{figure}[!th]
    \centering 
    \includegraphics[width=3in]{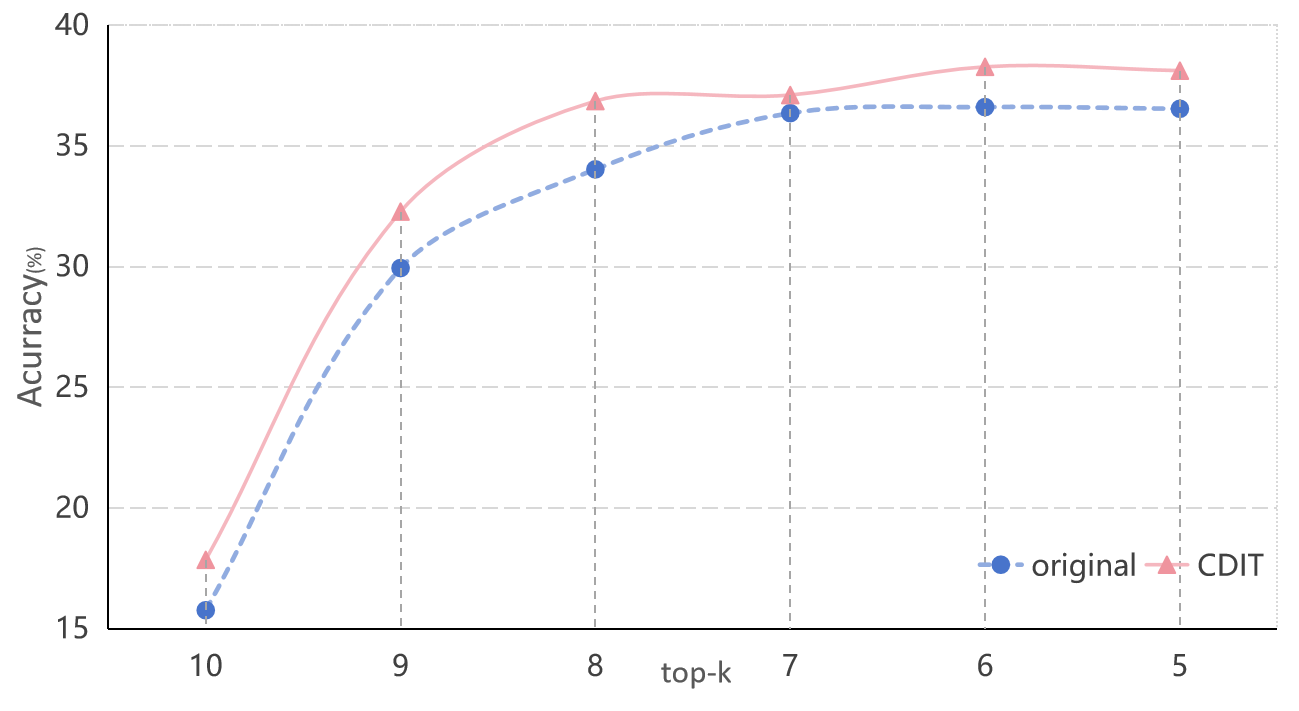}
    \caption{Top-k analysis on PopQA with Llama2-7b and IndexL2Flat index structure.}
    \label{fig:topk}
\end{figure}

\eat{Due to the fact that}\revise{As} CDIT is a method of pruning indexes and can be \eat{loosely}\revise{flexibly} \eat{coupled}\revise{integrated} with other \warn{RALMs} as a functional module,
it is actually an atomic component and is difficult to conduct ablation studies.
\revise{Thus, we analyze the impact of hyper-parameters and CMDs on CDIT.}
\eat{In order to analyze the impact of internal changes, we conducted tests about parameters of CDIT.}
Specific results are shown in \warn{Appendix~\ref{sec:ap-er}}.
\revise{\\ \textbf{Effects of \revise{varying} top-k.}
In order to analyze how the number of documents returned by the retriever affects the performance of CDIT,
we vary the number of retrieved documents (top-k) from 5 to 10 and test on PopQA dataset for all three vector indexes with llama2-7b as the generator.
As shown in Figure \ref{fig:topk}, CDIT shows improvements across various top-k, and the improvement is more significant under larger top-k.
The main reason for this is that larger top-k may return more useless information, which deteriorates the performance and can be filtered out by CDIT,
while smaller top-k provides little space for trimming where the performance gain is limited.}


\eat{We analyzed about how top-k affects model performance.
top-k means that k retrieved documents are passed to the language generator after retrieval.}

\eat{
A too large top-k may cause too much useless information, resulting in overall performance degradation, while too small one may result in little information to trim, such that the performance of CDIT is not significant.
\eat{Tests were conducted on PopQA for all three index structures using llama2-7b as the generator, with top-k ranking from 5 to 10.}

One possible reason is that CDIT can filter out more irrelevant information when retrieval information explodes.}

\begin{figure}[!th]
    \centering 
    \includegraphics[width=3in]{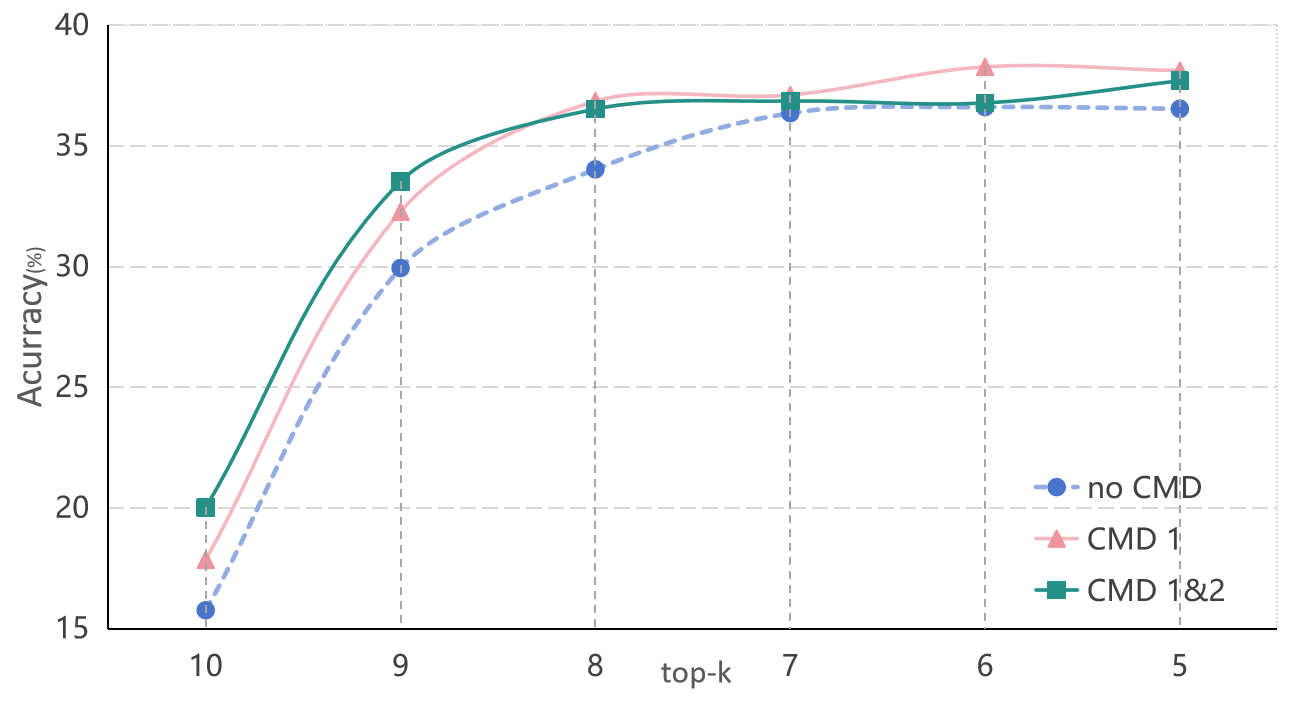}
    \caption{CMDs analysis on PopQA with Llama2-7b and IndexL2Flat index structure.}
    \label{fig:cmd}
\end{figure}

\begin{figure}[!th]
    \centering 
    \includegraphics[width=3in]{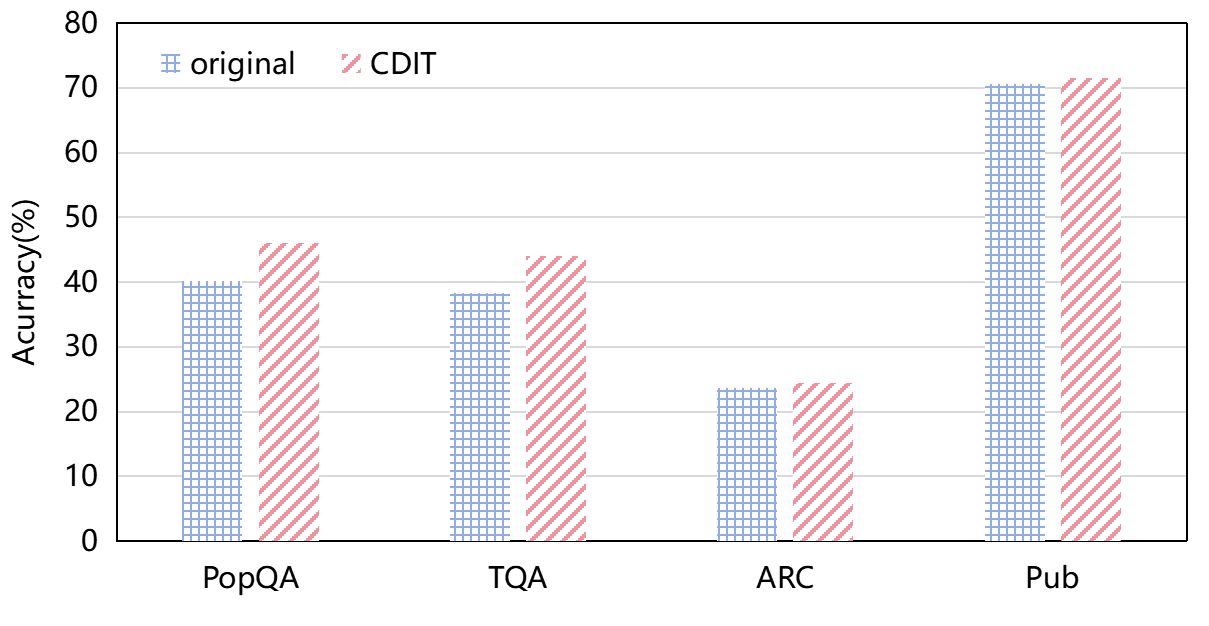}
    \caption{Self-RAG Integration analysis on IndexL2Flat with Llama2-7b.}
    \label{fig:selfrag}
\end{figure}

\revise{In particular, as for Llama3, a generator with strong language abilities,
We test its performance with a much larger top-k to evaluate how CDIT helps resolve the retrieval information explosion.
Previously, as shown in Table \ref{tab:result}, the performance gain of applying CDIT to Llama3 is limited,
where the possible reason may be that the strong language ability of Llama3 allows more contexts and better identifies irrelevant retrieval contexts independently without CDIT.
In order to investigate Llama3's ability limit of processing retrieval information explosion
and whether CDIT can still help to improve,
we vary (top-k) from 10 to 50 and test on PopQA with Llama3 as the generator.
As shown in Table \ref{tab:llama3}, the performance gain of CDIT is significantly enhanced as top-k increases,
which means that CDIT plays a better role when excessively large context information is fed to the LLM.}
\eat{In particular, for strong ability generators like Llama3, we use a much more larger top-k to test the retrieval information explosion.}
\eat{CDIT performed modestly on Llama3 in Table \ref{tab:result}.
We speculate that this may be due to the strong ability of Llama3, which allows it to accept more contexts and better identify irrelevant information independently.}
\eat{To address this issue, expansion experiments are conducted on PopQA with Llama3 being the language generator, inputting more retrieval documents into the LLM.
Results are shown in Table \ref{tab:llama3}.
It is apparent that the improvement brought by CDIT is significantly enhanced as the value of top-k increases. 
This means that when the information received is too large for LLM to handle, CDIT can play a better role.}
\begin{table} 
    \centering
    \scalebox{0.9}{
    \begin{tabular}{lccccc}
    \toprule
         Top-k& 10 & 20 & 30 & 40 & 50\\
         \midrule
         original& 42.20 & 41.03 & 41.20 & 39.86 & 16.51 \\
         CDIT& 42.37 & 41.25 & 41.45 & 40.95 & 24.35\\
         \hline
         impro&\textcolor{blue}{$\uparrow$ 0.17} &\textcolor{blue}{$\uparrow$ 0.22} &\textcolor{blue}{$\uparrow$ 0.25} &\textcolor{blue}{$\uparrow$ 1.09} &\textcolor{blue}{$\uparrow$ 7.84}\\
         \bottomrule
    \end{tabular}}
    \caption{The performance of CDIT on Llama3 as top-k ranks from 10 to 50.}
    \label{tab:llama3}
\end{table}
\revise{\\ \textbf{Effects of CMDs.}
The CMD $\phi_1$ does not describe all constraints of retrieved data.
In order to specify the constraints more accurately,
we need to consider the relationship between other components of the sentences.
For example, attributives and adverbials also constraint the consistency of the sentences,
and the corresponding CMD can be written as below:
\begin{equation*}
\label{eq:md_2}
    \begin{aligned}
    \phi_2:
&s_1[sid] \sim s_2[sid] \rightarrow   \\
&s_1[att] \approx s_2[att] \land s_1[adv] \approx s_2[adv]
\end{aligned}
\end{equation*}
where $att, adv$ denote the attributive and adverbial of the sentence.
We add CMD $\phi_2$ to the comparing step,
such that the consistency of $sid$ requires simultaneous satisfaction of both CMD $\phi_1$ and CMD $\phi_2$.
As shown in Figure \ref{fig:cmd}, different CMDs have an influence on the accuracy of CDIT.
Thus, it is desirable to investigate the optimized combinations of various CMDs in future.}
\revise{\\ \textbf{Integration with other RALMs.}
We analyze the performance of CDIT when integrated with enhanced RAG models.
As CDIT directly modifies the indexes of database, it has strong flexibility and can be easily integrated with
existing RAG models to improve the answer quality.
We integrate CDIT with Self-RAG\cite{asai2023selfrag}, which is a refined RAG model by improving knowledge retrieval,
and evaluate the accuracy of the answers.
Llama2-7b is used in this test, and other settings are unchanged.
As shown in Figure \ref{fig:selfrag}, CDIT consistently improves the accuracy of answers
after integrating with self-rag, where the average increase is 3.62\%.
This shows that based on existing state-of-the-art RAG models that mainly improve 12.3\%,
CDIT could further enhance the performance by refining data quality.}
\\ \textbf{LLM costs.}
We analyze the costs of utilizing LLMs in CDIT, which has been validated acceptable.
Time to call LLMs is considered to assess the cost with results as follows.
As shown in Table \ref{tab:time}, even with the maximum number of calls to LLMs, without considering repeated queries, the time did not exceed 6 minutes. 
Compared to the time required for the inference of language model itself, we consider this cost to be reasonable.
\begin{table}[!th]
    \centering
    \begin{tabular}{ccccc}
    \toprule
        Dataset & PopQA & TQA & ARC & Pub \\
        \midrule
        Time(s) & 59.94 & 351.05 & 58.62 & 29.61\\
    \bottomrule
    \end{tabular}
    \caption{Time to call LLMs in CDIT.}
    \label{tab:time}
\end{table}
\\ \textbf{Case Study.}
We use the scenario depicted in Figure \ref{fig:framework_intro} as an example to explain the working principle of CDIT. 
In this scenario, the input query is "\textit{Who turned on the radio?}", and the two relevant retrieved contexts are: \textcircled{\small{1}} "\textit{Mary turned off the radio.}" \textcircled{\small{2}} "\textit{Jack turned on the radio.}"

Before CDIT, the basic RALMs will answer "\textit{Mary}." due to the first retrieved context, while they actually give a wrong answer.
For CDIT, First it determines that the verbs "\textit{turn on}" in query and "\textit{turn off}" in context \textcircled{\small{1}} are contradictory, hence these two sentences are inconsistent.
As a result, context \textcircled{\small{1}} is discarded.
At this point, the language model only receives query and context \textcircled{\small{2}}.
According to context \textcircled{\small{2}}, RALMs with CDIT can answer "\textit{Jack}." correctly.

\eat{combination effects of CDIT with other methods.
CDIT directly modifies the indexes of database, thus it has strong versatility and can be well combined with other methods to improve the quality of answers.}
\eat{We try to integrate CDIT on self-rag\cite{asai2023selfrag} framework, different from the standard RAG models\cite{Guu2020REALMRL} used in the above experiments.}
\eat{Llama2-7b are used in the test with other settings unchanged.}
\eat{From the results shown in Figure \ref{fig:topk}(b), we can conclude that CDIT can be combined with popular RAG methods easily and effectively.}
\eat{After other methods had improved the accuracy of answers, CDIT could further enhance the effects by refining data quality.}

\section{Conclusion}
\eat{
\revise{We propose Context Matching Dependency (CMD) and the Context-Driven Index Trimming (CDIT) framework to enhance the precision of RAG from the perspective of improving data quality by modifying vector database indexes.
CDIT achieves this by:
(1) leveraging LLMs to extract linguistic components and determine their similarity;
(2) using CMDs to specify consistency constraints over data which assess the similarity of $sid$s by logical rules;
(3) a pruning algorithm that modifies and updates database indexes.}
\warn{Experiments succeeded on Faiss and increased accuracy by 3.79\% compared to original models on average, validating the effectiveness of CDIT and demonstrating that it can be integrated with various RALMs to achieve better results.}
}

Our study presents a Context-Driven Index Trimming (CDIT) framework, which enhances the accuracy of RALMs by focusing on data quality within vector databases. 
Experiments show an average 3.76\% increase in accuracy, highlighting the robustness of CDIT  across models and indexing methods. 
While challenges in long text handling and reliance on LLMs are noted, the adaptability and potential of CDIT suggest a bright future in NLP.

\section*{Limitations}
Limitations still exist in our work.
Firstly, long texts may cause a subpar performance of CDIT.
A possible reason is the complexity of long texts, making it difficult to extract and compare basic semantic components.
What's more, the basic components mentioned above may be incompetent to represent long text, resulting in error judgements.
Secondly, the CMDs in this article are proposed manually based on our experience, so they may not be thorough and accurate enough to describe all the constraints of retrieved data.
Regarding the two limitations, we plan to conduct further research on the mining of CMDs in the future to enable it to represent the constraints of various types of text more accurately.
Finally, over-reliance on GPT is another potential problem. 
The extraction and comparison of components need online LLMs, which may be a trouble for a completely offline environment with considerable costs.
We may subsequently consider employing proper lexical analyzers such as dependency parsing\cite{Manning2014TheSC}, Stanza\cite{qi2020stanza}, etc, to mitigate this issue.

\bibliography{custom}
\appendix
\section{Implementation}
\label{sec:ap-ip}
\subsection{Datasets Details}
For all the datasets, we set the maximum new token number to 100 tokens.
For PopQA and ARC-Challenge, we retrieve top-k documents from the 2018 English Wikipedia.
For TriviaQA, we additionally retrieve documents using Google Programmable Search.
\subsection{Query Rewriting.}
\label{sec:ap-qr}
After the indexes have been trimmed, we rewrite the query to test the performance.
We follow the approach of \citeauthor{Feng2023SynergisticIB}, combining the original query and the top-1 retrieval document to form a new query.
Therefore, the new query sent to the language generator is shown in Table \ref{tab:prompt-qr}.

\subsection{Configuration for self-rag}
We follow the method of \citeauthor{asai2023selfrag}, employing the pre-trained weights\footnote{\url{https://huggingface.co/selfrag/selfrag_llama2_7b}}.
The integration experiment uses CDIT first to modify indexes, then self-rag is employed to test the overall quality.
The default configuration is employed for everything else.

\section{Prompting}
Based on time and performance considerations, we choose OpenAI\footnote{\url{https://platform.openai.com/docs/api-reference}} GPT-3.5-turbo API as the employed LLM.
CDIT primarily utilizes LLM in three parts, with detailed prompts as follows.
\subsection{Extraction and Comparison.}
\label{sec:ap-e&c}
Prompting is used in GPT-3.5 to achieve the functionality of extracting components and utilizing the CMDs for comparison.
Prompt 1 and 2 in Table \ref{tab:prompt-ec} are utilized respectively for trimming with CMD $\phi_1$ and CMD $\phi_1$ \& $\phi_2$.

\subsection{Prompt for Answers}
\label{sec:ap-ti}
After retrieval, we combine the retrieval contexts and other instructions to prompt the language generator for the final answers.
For ARC-Challenge, we follow \citeauthor{asai2023selfrag}, designing task instructions shown in Table \ref{tab:prompt-ins}.
For other tasks, we do not design additional task instructions.
The final prompts are shown in Table \ref{tab:prompt-all}.

\section{Experiment Results}
\label{sec:ap-er}
Table \ref{tab:topk} shows the specific results of the top-k experiment.
In this experiment, Llama2-7b, IndexL2Flat are employed as the language generator and the indexing structure.
Top-k varies from 10 to 5. 
Table \ref{tab:cmd-r} shows the results of the cmd expended experiment.
In this experiment, Llama2-7b, IndexL2Flat are employed as the language generator and the indexing structure.
CMD $\phi_1, \phi_2$ are employed differently.
Table \ref{tab:selfrag} shows the results of the integration experiment between CDIT and Self-RAG\cite{asai2023selfrag}.
CDIT is employed firstly to enhance the data quality of retrieved contexts.
Then, the model was trained and tested using the Self-RAG approach.
Llama2-7b is employed as the language generator in this experiment, and top-k is 10.

\begin{table*}[!ht]
\centering
\begin{tabular}{p{450pt}}
\toprule
\textbf{\textcolor{blue}{Original Query:}} What is Henry Feilden's occupation?\\
\midrule
\textbf{\textcolor{blue}{Retrieved Documents:}} \\
$[1]$ Henry Feilden (Conservative politician) Henry Master Feilden (21 February 1818 – 5 September 1875) was an English Conservative Party politician. \\
$[2]$ Henry Wemyss Feilden Colonel. Henry Wemyss Feilden, CB (6 October 1838 – 8 June 1921) was a British Army officer, Arctic explorer and naturalist. \\
\midrule
\textbf{New Query Structure:} \\
Given a question [\textcolor{blue}{\underline{original query}}] and its possible answering passages [\textcolor{blue}{\underline{top-1 retrieved documents}}], Now give a possible answer. \\
\midrule
\textbf{New Query:} \\
Given a question [\textcolor{blue}{What is Henry Feilden's occupation?}] and its possible answering passages [\textcolor{blue}{Henry Feilden (Conservative politician) Henry Master Feilden (21 February 1818 – 5 September 1875) was an English Conservative Party politician.}], Give a possible answer. \\
\bottomrule
\end{tabular}
\caption{Construction of new queries in query rewriting}
\label{tab:prompt-qr}   
\end{table*}

\begin{table*}[!ht]
\centering
\begin{tabular}{p{450pt}}
\toprule
\textbf{CMD:} CMD $\phi_1$\\
\hline
\textbf{Prompt 1:}
    You are a cautious language assistant. \\
    \#\#\#\textcolor{blue}{[Rules]} Here are some language rules: \\
    \# If the two sentences can be identified as similar, then the subjects, predicates and objects of the two sentences are similar. Be especially mindful of predicate phrases that appear similar but actually have opposite meanings, which make sentences dissimilar. \\
    \#\#\#\textcolor{blue}{[Instructions]} Are the following statements similar with the question? Just say True if they are; otherwise just say False. Only output one word. \\
\hline
\textbf{Sentences:} \\
He \textcolor{red}{\underline{turned on}} the radio.\\
He \textcolor{red}{\underline{turned off}} the radio.\\
\hline
\textbf{Answer:} False. \textcolor{green}{\usym{2714}} \\
\bottomrule
\toprule
\textbf{CMD:} CMD $\phi_1$\&$\phi_2$\\
\hline
\textbf{Prompt 2:}
    You are a cautious language assistant. \\
    \#\#\#\textcolor{blue}{[Rules]} Here are some language rules: \\
    \# If the two sentences can be identified as similar, then the subjects, verbs and objects of the two sentences are similar. Be especially mindful of verb phrases that appear similar but actually have opposite meanings, which make sentences dissimilar. \\
    \# If the two sentences can be identified as similar, then the adverbials and attributives of the two sentences are similar. \\
    \#\#\#\textcolor{blue}{[Instructions]} Are the following statements similar with the question? Just say True if they are; otherwise, just say False. Only output one word. \\
\hline
\textbf{Sentences:} \\
He turned on the radio \textcolor{red}{\underline{at five}}.\\
He turned on the radio \textcolor{red}{\underline{at six}}.\\
\hline
\textbf{Answer:} False. \textcolor{green}{\usym{2714}}\\
\bottomrule
\end{tabular}
\caption{Prompts used in extraction and comparison}
\label{tab:prompt-ec}   
\end{table*}

\begin{table*}[!th]
\centering
\begin{tabular}{p{450pt}}
\toprule
\textbf{\textcolor{blue}{Query:}} What is Henry Feilden's occupation? \\
\midrule
\textbf{\textcolor{blue}{Retrieved Documents:}}\\
$[1]$ Henry Feilden (Conservative politician) Henry Master Feilden (21 February 1818 – 5 September 1875) was an English Conservative Party politician. \\
$[2]$ Henry Wemyss Feilden Colonel Henry Wemyss Feilden, CB (6 October 1838 – 8 June 1921) was a British Army officer, Arctic explorer and naturalist. \\
\midrule
\textbf{Prompt Structure:} \\
\#\#\#Background: \{\textcolor{blue}{\underline{retrieved documents}}\} \\
\#\#\#Instruction: \{\textcolor{blue}{\underline{query}}+\textcolor{blue}{\underline{task instructions}}\} \\
\#\#\#Response: \\
\midrule
\textbf{Prompt:}\\
\#\#\#Background: \{\textcolor{blue}{$[1]$ Henry Feilden (Conservative politician) Henry Master Feilden (21 February 1818 – 5 September 1875) was an English Conservative Party politician.
$[2]$ Henry Wemyss Feilden Colonel Henry Wemyss Feilden, CB (6 October 1838 – 8 June 1921) was a British Army officer, Arctic explorer and naturalist.}\} \\
\#\#\#Instruction: \{\textcolor{blue}{What is Henry Feilden's occupation?}\} \\
\#\#\#Response:\\
\bottomrule
\end{tabular}
\caption{Prompts for generating answers.}
\label{tab:prompt-all} 
\end{table*}

\begin{table*}[!ht]
\centering
\begin{tabular}{p{450pt}}
\toprule
\textbf{Task Instruction}\\
\midrule
Given four answer candidates, A, B, C and D, choose the best answer choice. Please answer with the capitalized alphabet only, without adding any extra phrase or period. \\
\bottomrule
\end{tabular}
\caption{Task instruction for ARC-Challenge.}
\label{tab:prompt-ins} 
\end{table*}

\begin{table*}[!ht]
    \centering
    \begin{tabular}{l|c|ccccccc}
    \toprule
        \multirow{2}{*}{Index} & \multirow{2}{*}{Method} & \multicolumn{6}{c}{\textbf{Top-k}} &\multirow{2}{*}{\textbf{Avg-Impro}} \\
        & & 10 & 9 & 8 & 7 & 6 & 5\\
        \midrule
        \multirow{3}{*}{IndexL2Flat} & original & 15.76 & 29.94 & 34.03 & 36.36 & 36.61 & 36.54  &\multirow{3}{*}{\textcolor{blue}{$\uparrow$ 1.88}}\\
         & CDIT & \textbf{17.85} & \textbf{32.28} & \textbf{36.86} & \textbf{37.11} & \textbf{38.28} & \textbf{38.12}\\
         & \textit{Impro.} & \textcolor{blue}{$\uparrow$ 2.09} & \textcolor{blue}{$\uparrow$ 2.34} & \textcolor{blue}{$\uparrow$ 2.83} & \textcolor{blue}{$\uparrow$ 0.75} & \textcolor{blue}{$\uparrow$ 1.67} & \textcolor{blue}{$\uparrow$ 1.58} \\
         \midrule
        \multirow{3}{*}{IndexHNSWFlat} & original & 19.52 & 30.52 & 34.61 & 34.52 & 35.69 & 34.19 &\multirow{3}{*}{\textcolor{blue}{$\uparrow$ 2.84}}\\
         & CDIT & \textbf{27.61} & \textbf{34.62} & \textbf{36.03} &\textbf{36.61} & \textbf{36.11} & \textbf{35.11} \\
         &\textit{Impro.} &\textcolor{blue}{$\uparrow$ 7.09} &\textcolor{blue}{$\uparrow$ 4.10} &\textcolor{blue}{$\uparrow$ 1.42} &\textcolor{blue}{$\uparrow$ 1.09} &\textcolor{blue}{$\uparrow$ 0.42} &\textcolor{blue}{$\uparrow$ 0.92}\\
         \midrule
         \multirow{3}{*}{IndexIVFFlat} & original & 20.26 & 31.35 & 34.78 & 35.94 & \textbf{36.86} & \textbf{37.78} &\multirow{3}{*}{\textcolor{blue}{$\uparrow$ 1.85}}\\
         & CDIT & \textbf{25.44} & \textbf{35.03} & \textbf{35.94} & \textbf{37.86} & 36.20 & 37.61\\
         &\textit{Impro.} &\textcolor{blue}{$\uparrow$ 5.18} &\textcolor{blue}{$\uparrow$ 3.68} &\textcolor{blue}{$\uparrow$ 1.16} &\textcolor{blue}{$\uparrow$ 1.92} &$\downarrow$ 0.66 &$\downarrow$ 0.17\\ 
         \bottomrule
    \end{tabular}
    \caption{Changes in accuracy for models with and without CDIT on PopQA as the top-k parameter varies.}
    \label{tab:topk}
\end{table*}

\begin{table*}[!ht]
    \centering
    \scalebox{1.0}{
    \begin{tabular}{l|cccccc}
    \toprule
        top-k & 10 & 9 & 8 & 7 & 6 & 5\\
        \midrule
        no CMD & 15.76 & 29.94 & 34.03 & 36.36 & 36.61 & 36.54\\
        \midrule
        CMD $\phi_1$ & 17.85 & 32.28 & 36.86 & 37.11 & 38.28 & 38.12\\
        CMD $\phi_1$\&$\phi_2$ & 20.02 & 33.53 & 36.53 & 36.86 & 36.78 & 37.70\\
        \bottomrule
    \end{tabular}
    }
    \caption{Accuracy of CDIT with different CMDs.}
    \label{tab:cmd-r}
\end{table*}

\begin{table*}[!ht]
    \centering
    \begin{tabular}{l|c|lllll}
    \toprule
        \textbf{Index} & \textbf{Method} & \textbf{PopQA} & \textbf{TQA} & \textbf{ARC} &\textbf{Pub} & \textit{Avg.}\\
        \midrule
        \multirow{2}{*}{IndexL2Flat} & original & 40.12 & 38.30 & 23.63 & 70.62 & 43.17\\
         & CDIT & \textbf{46.04} & \textbf{44.02} & \textbf{24.40} &\textbf{71.53}  
  &\textbf{46.50} \footnotesize{\textcolor{blue}{$\uparrow$ 3.33}}\\
         \midrule
         \multirow{2}{*}{IndexHNSWFlat} & original & 41.11 & 38.29 & 22.36 & 70.72 & 43.37\\
         & CDIT & \textbf{43.54} & \textbf{43.80} & \textbf{24.91} &\textbf{72.68} & \textbf{46.24} \footnotesize{\textcolor{blue}{$\uparrow$ 2.87}}\\
         \midrule
         \multirow{2}{*}{IndexIVFFlat} & original& 40.45 & 38.82 & 24.48 & 70.72 & 43.62\\
         & CDIT & \textbf{44.04} & \textbf{44.04} & \textbf{26.37} & 72.34 & \textbf{46.70} \footnotesize{\textcolor{blue}{$\uparrow$ 3.08}}\\
    \bottomrule
    \end{tabular}
    \caption{The performance of CDIT integrated with self-rag on three datasets with top-k being 10.}
    \label{tab:selfrag}
\end{table*}

\begin{figure*}[!b]
\centering
\subfloat[Top-k Effects on IndexL2Flat.]{
		\includegraphics[width=3in,height=1.8in]{picture/fig_topk_l2.png}}
\subfloat[Integration Effects on IndexL2Flat.]{
		\includegraphics[width=3in,height=1.8in]{picture/fig_selfrag_l2.png}}
\\
\subfloat[Top-k Effects on IndexHNSWFlat.]{
		\includegraphics[width=3in,height=1.8in]{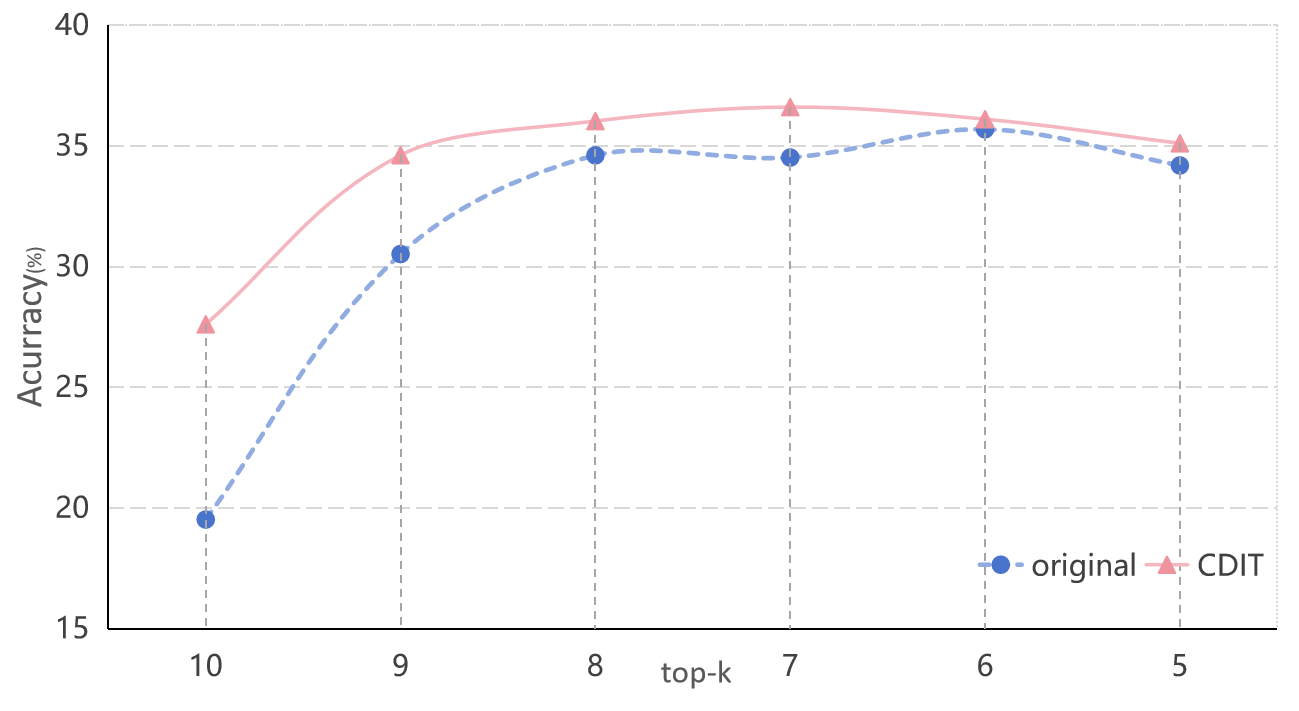}}
\subfloat[Integration Effects on IndexHNSWFlat.]{
		\includegraphics[width=3in,height=1.8in]{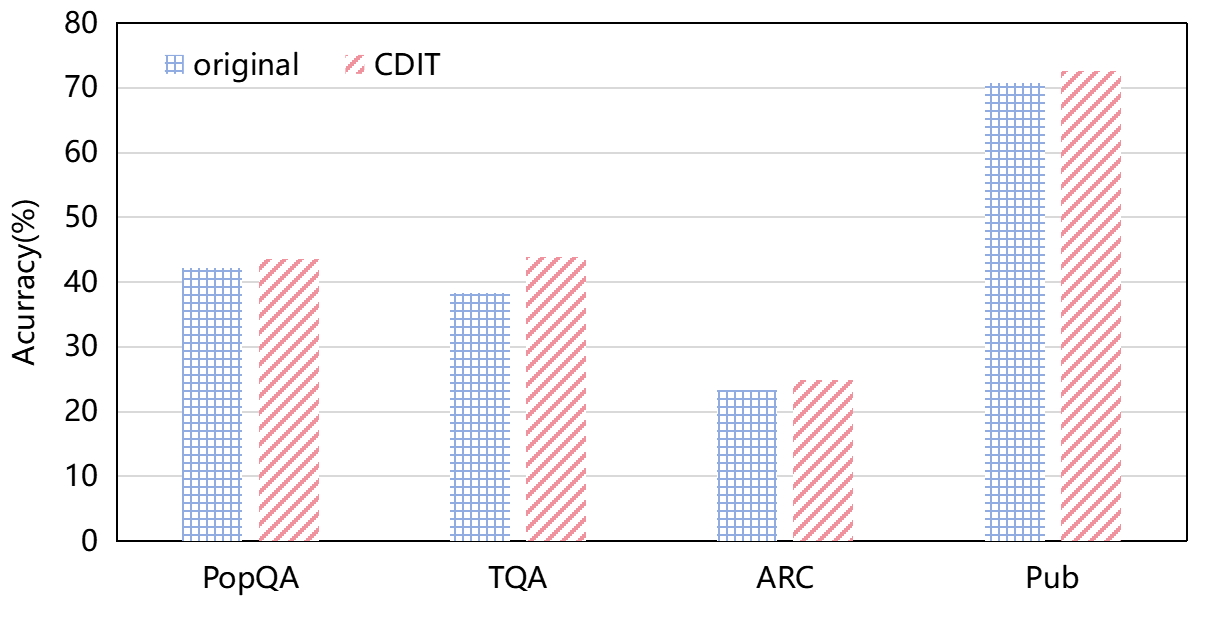}}
\\
\subfloat[Top-k Effects on IndexIVFFlat.]{
		\includegraphics[width=3in,height=1.8in]{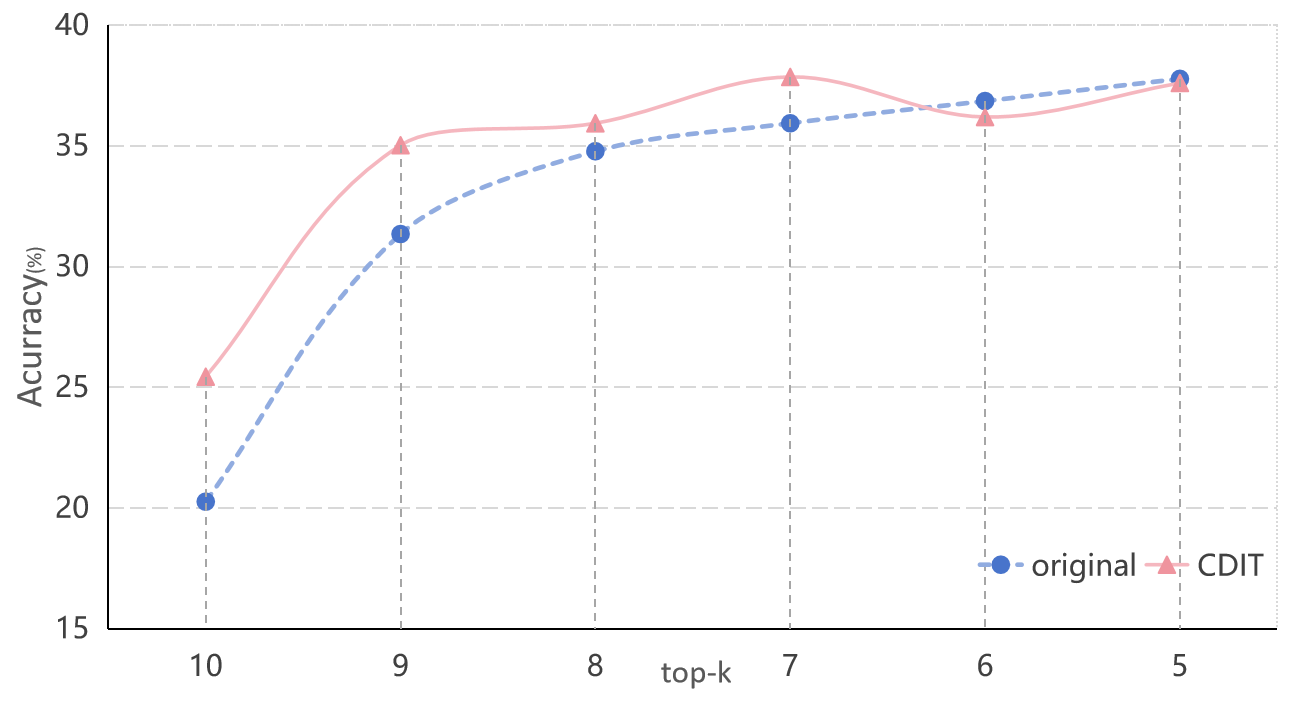}}
\subfloat[Integration Effects on IndexIVFFlat.]{
		\includegraphics[width=3in,height=1.8in]{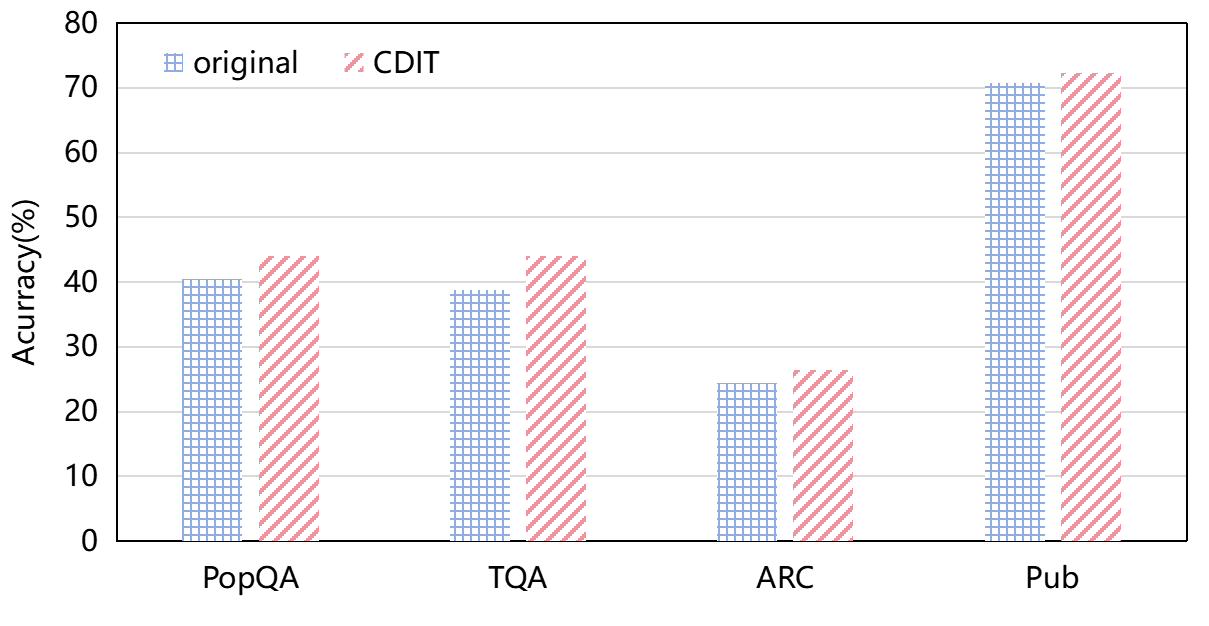}}
\caption{Additional experiments of top-k and integration with Self-RAG on Llama2-7b.}
\label{fig_6}
\end{figure*}

\end{document}